\documentclass[10.5pt]{article}
\usepackage[margin=0.8in]{geometry}

\usepackage{amsmath,amssymb}
\usepackage{tikz-cd}
\usepackage{multicol}
\usepackage{booktabs}
\usepackage{graphicx}
\usepackage{caption}
\usepackage{makecell}
\usepackage{multirow}
\usepackage[table]{xcolor}
\usepackage{svg}
\usepackage{hyperref}

\newcommand{\better}[1]{\textcolor{green!50!black}{#1}}
\newcommand{\worse}[1]{\textcolor{red!75!black}{#1}}

\newcommand{\diffcell}[2]{\makecell{#1\\{\footnotesize #2}}}
\newcommand{\lowfreq}[1]{\textcolor{cyan!55!blue}{#1}}
\newcommand{\highfreq}[1]{\textcolor{violet!75!black}{#1}}

\title{\textbf{Rethinking One-Step Image Editing through ChordEdit: \\ Reproduction, Simplification, and New Insights}}
\author{Minghan Li, Jeremy Moebel, and Mengyu Wang\\ Harvard AI and Robotics Lab}
\date{\today}

\begin{document}
\maketitle

\begin{center}
\vspace{-2mm}
\includegraphics[width=0.49\textwidth]{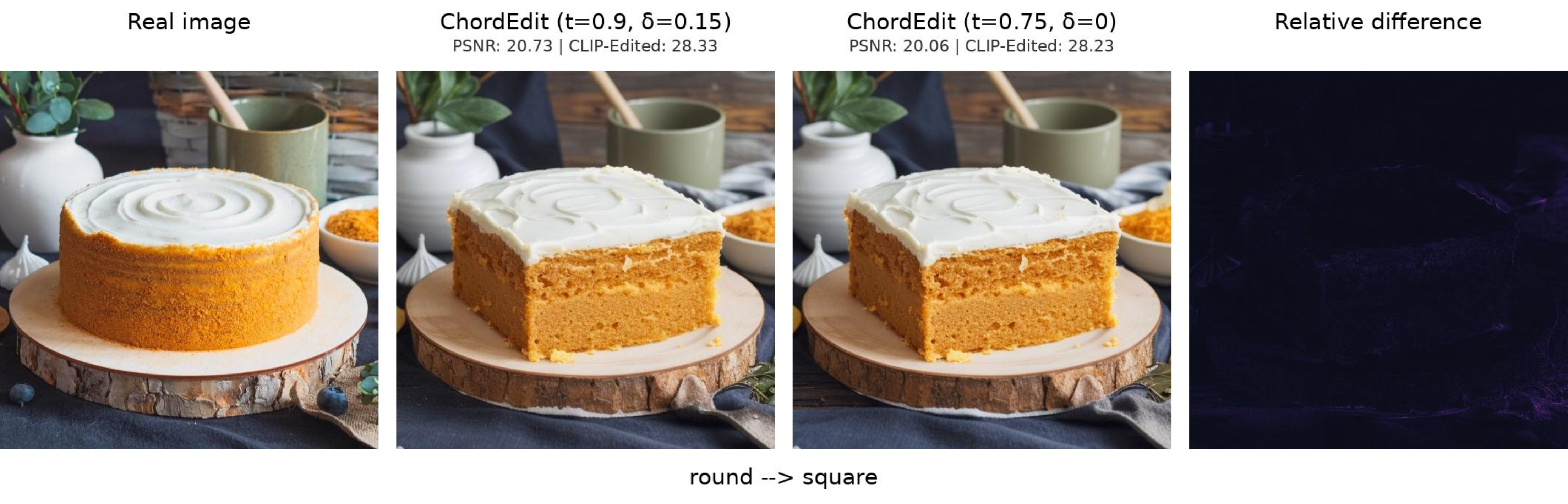}
\vspace{0.3mm}
\includegraphics[width=0.49\textwidth]{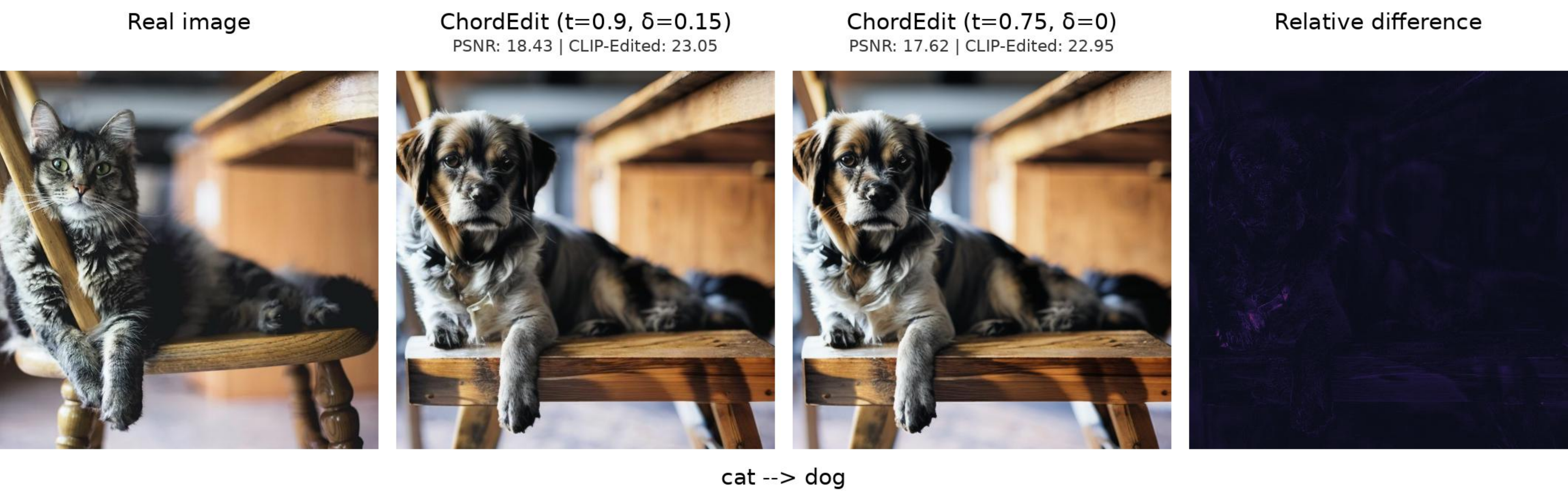}
\vspace{0.3mm}
\includegraphics[width=0.49\textwidth]{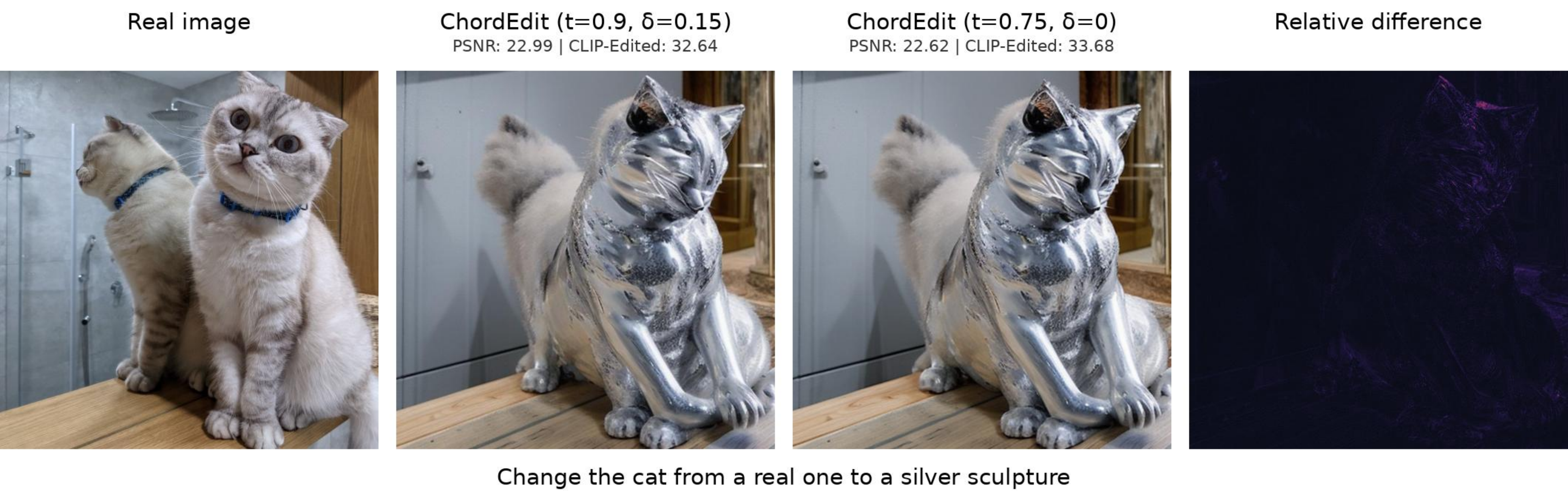}
\vspace{0.3mm}
\includegraphics[width=0.49\textwidth]{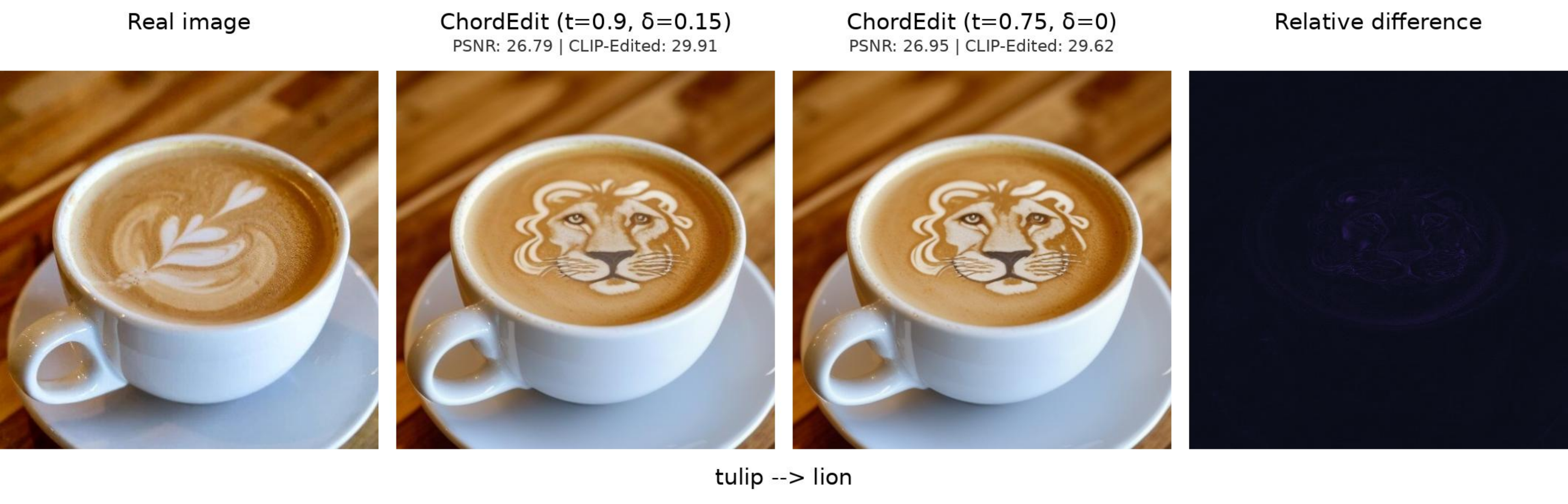}
\vspace{0.3mm}
\includegraphics[width=0.49\textwidth]{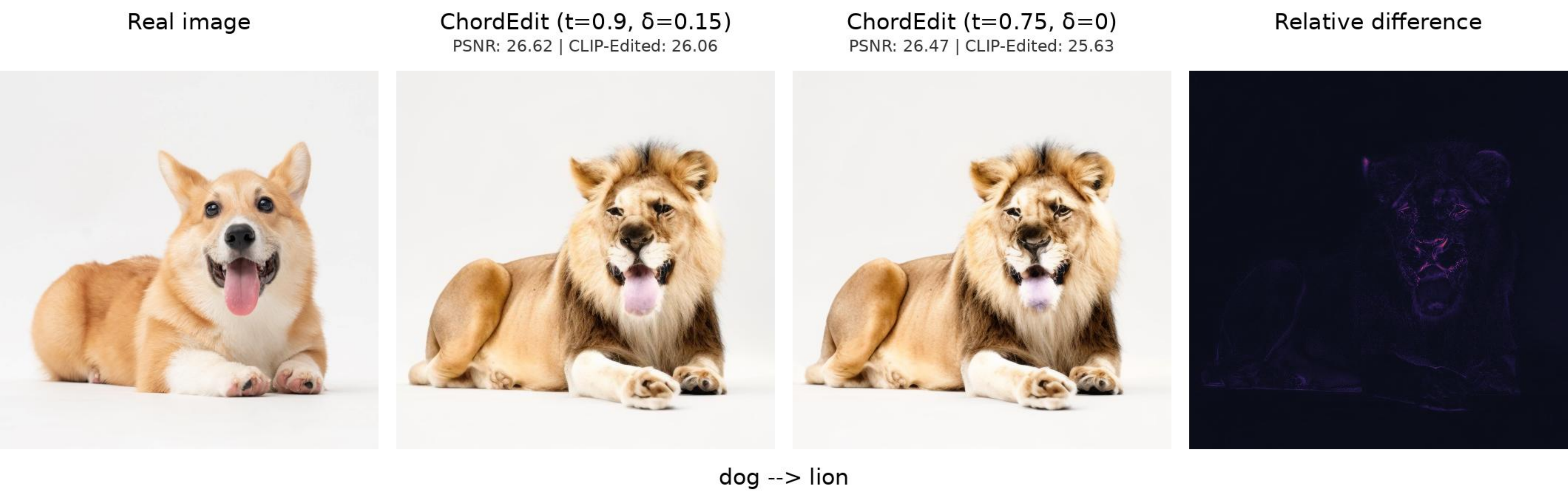}
\vspace{0.3mm}
\includegraphics[width=0.49\textwidth]{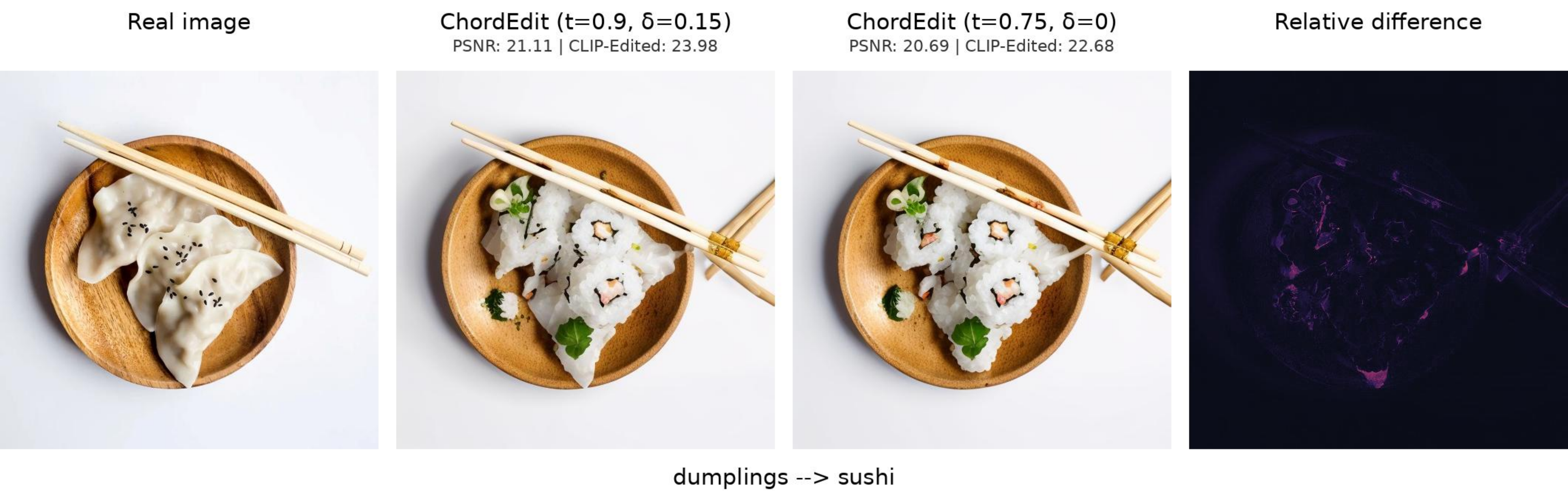}
\vspace{0.3mm}
\includegraphics[width=0.49\textwidth]{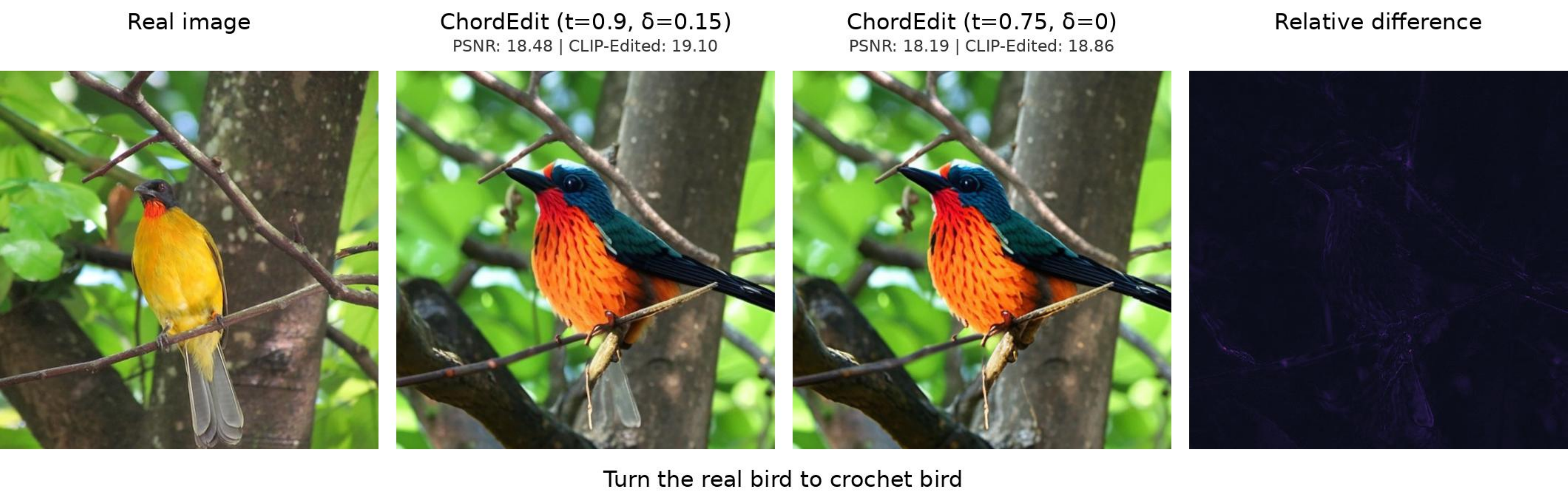}
\vspace{0.3mm}
\includegraphics[width=0.49\textwidth]{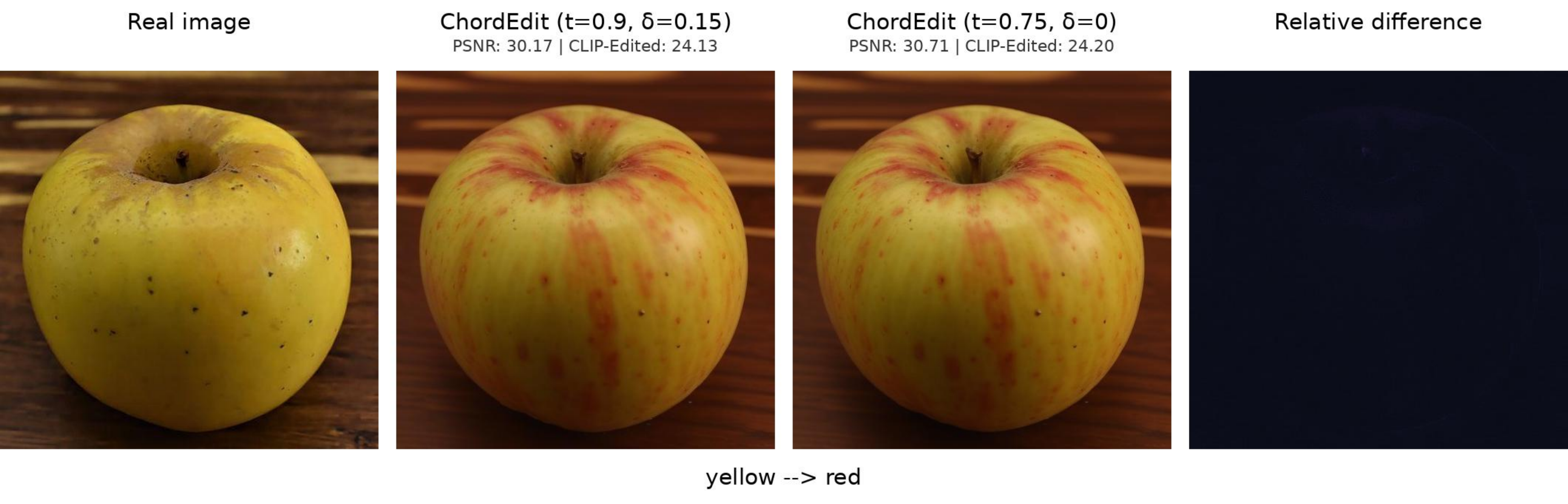}

\vspace{-3mm}
\captionof{figure}{\small Main takeaway: the default ChordEdit update ($t=0.9,\delta=0.15$) is visually almost equivalent to directly editing at its dominant effective timestep ($t=0.75,\delta=0$). These examples suggest that ChordEdit's gains are driven primarily by choosing the right editing timestep, rather than by the full chord-transport window itself.}
\label{fig:default-vs-t075-anchor-qualitative}
\vspace{-2mm}
\end{center}

\begin{abstract}
\noindent One-step image editing is important for making text-guided editing fast, practical, and easy to deploy, but its underlying mechanism is still not fully understood. We revisit ChordEdit through reproduction, ablation, and simplification. Our analysis shows that \textbf{a)} the chord window $\delta$ largely acts as an effective timestep shift from \lowfreq{$t$} to \lowfreq{$t - \delta$}; \textbf{b)} chord transport acts on high-noise images and mainly performs low-frequency semantic editing; and \textbf{c)} proximal alignment acts on low-noise images and complements it by adding high-frequency target details. In this view, ChordEdit naturally decomposes editing into a coarse low-frequency transport stage and a fine high-frequency alignment stage. These findings suggest a path toward prompt-conditioned dynamic timestep selection for adaptive image editing. All code and results can be found at \href{https://github.com/Harvard-AI-and-Robotics-Lab/ChordEdit-Reproduction}{link}. 
\end{abstract}

\section{Reformulated ChordEdit}

To make the method as clear and concise as possible, we follow the original ChordEdit~\cite{lu2026chordedit} paper and distill its method section into a simplified presentation. Most of the notation is kept consistent with the original formulation, while a few symbols are refined to reduce ambiguity in our rectified-flow implementation.

\textbf{Chord Control Field.} We first recall the practical Chord Control Field derived in Section~4.2 of the original paper~\cite{lu2026chordedit}. This field is the key quantity used to transfer the chord-based update from the source image $x_{src}$ to the target image $x_{tar}$: 
\begin{equation}
    \hat{u}_t(x_t) = \frac{t\mathbf{R}(x_{t- \delta}, t - \delta) + \delta\mathbf{R}(x_t, t)}{t + \delta}.
\end{equation}
Here, $x_t$ denotes the source image with the noise level at timestep $t$, and $\delta$ specifies the time window used to estimate the local chord direction. $\mathbf{R}(x_t, t)$ is the observable proxy field. For flow-based models such as SD-Turbo~\cite{sauer2023add}, the proxy field $\mathbf{R}(x_t, t)$ is instantiated as the conditional velocity difference $\Delta \mathbf{v}(x_t, t)= \mathbf{v}(x_t, t, c_{\mathrm{tar}}) - \mathbf{v}(x_t, t, c_{\mathrm{src}})$. Substituting this quantity into the chord control field yields the following numerical update:
\begin{equation}\label{eq:chord-control-field-rf}
    \hat{u}_t(x_t) = \frac{t \Delta \mathbf{v}(x_{t-\delta}, t - \delta) + \delta \Delta\mathbf{v}(x_t, t)}{t + \delta}.
\end{equation}
This estimate gives the editing direction at the current noisy source state. The one-step edited prediction is then obtained by moving $x_{src}$ along this direction with step scale $\lambda = 1$:
\begin{equation}
    \text{\lowfreq{One-step:}} \quad \lowfreq{x^*_{tar} = x_{src} + \lambda \hat{u}_t(x_t).}
\end{equation}
\textbf{Proximal Refinement.} Section~4.3 of the original paper~\cite{lu2026chordedit} further introduces \textit{Proximal Refinement} to improve semantic alignment of the one-step edited prediction $x^*_{tar}$. To make this step concrete, we take rectified flow models as an example. Under the rectified flow parameterization, a noisy sample at timestep $t$ can be written as a linear interpolation between the clean image and Gaussian noise. Therefore, proximal refinement first maps the edited prediction back to a noisy state:
\begin{equation}
    x^*_{t}
    = (1 - t) x^*_{tar}
    + t \epsilon,
    \qquad \epsilon \sim \mathcal{N}(0, I).
\end{equation}
Starting from this noisy state, we then perform one additional target-conditioned forward step. Since the velocity field $\mathbf{v}(x^*_{t}, t, c_{\mathrm{tar}})$ points from the noisy state toward the target-prompt image, the refined output is obtained as
\begin{equation}
    \text{\highfreq{Two-step:}} \quad \highfreq{x^{**}_{\mathrm{tar}}
    = x^*_{t} + t \mathbf{v}\left(x^*_{t}, t, c_{\mathrm{tar}}\right).}
\end{equation}
This second pass encourages the edited result to better match the target prompt while remaining close to the one-step prediction. 

\textbf{Algorithm.} The following algorithm summarizes how the Chord Control Field and Proximal Refinement are combined to get the final target images. To avoid ambiguity, we use \highfreq{$t^{**}$} to denote the proximal refinement time, while \lowfreq{$t$} denotes the step time used in the Chord Control Field.

\begin{center}
\begin{minipage}{0.65\textwidth}
\hrule
\vspace{0.4em}
\textbf{Algorithm 1 ChordEdit Algorithm (based on Rectified Flows) }
\vspace{0.4em}
\hrule
\vspace{0.6em}
{\renewcommand{\arraystretch}{1.1}%
\begin{tabular}{@{}r l@{}}
1: & \textbf{Inputs:} source image $x_{src}$; prompts $c_{src}, c_{tar}$; step scale $\lambda=1$;  \\
   & \qquad \qquad \lowfreq{step time $t$}; window $\delta$; \highfreq{proximal refinement time $t^{**}$}. \\
2: & \textbf{Output:} edited image $\lowfreq{x_{tar}^{*}}$ or $\highfreq{x_{tar}^{**}}$ \\
3: & \textbf{Init:} sample noise $\epsilon \sim \mathcal{N}(0,I)$. \\
\rowcolor{cyan!8}
4: & \textbf{Chord Control Field:} \\
\rowcolor{cyan!8}
   & \qquad $\displaystyle x_s=(1-s)x_{src}+s\epsilon$  \quad $\Delta\mathbf{v}(x_s,s)=\mathbf{v}(x_s,s,c_{tar})-\mathbf{v}(x_s,s,c_{src})$,  \quad $\forall s\in\{t,t-\delta\}$ \\
\rowcolor{cyan!8}
   & \qquad $\displaystyle \hat{u}_t(x_t) \leftarrow
      \frac{t\Delta\mathbf{v}(x_{t-\delta}, t-\delta)
      + \delta\Delta\mathbf{v}(x_t,t)}{t+\delta}$. \\
\rowcolor{cyan!8}
   &\qquad  $\displaystyle \lowfreq{x_{tar}^{*} \leftarrow x_{src} + \lambda \hat{u}_t(x_t)}$ \\
\rowcolor{violet!8}
5: & \textbf{Proximal Refinement (Optional):} \\ 
\rowcolor{violet!8}
   & \qquad $\displaystyle x_{t^{**}}^{*} \leftarrow (1-t^{**})x_{tar}^{*}+t^{**}\epsilon$ \\
\rowcolor{violet!8}
   & \qquad $\displaystyle \highfreq{x_{tar}^{**} \leftarrow x_{t^{**}}^{*}
      + t^{**}\mathbf{v}(x_{t^{**}}^{*}, t^{**}, c_{tar})}$ \\
6: & \textbf{Return} $\lowfreq{x_{tar}^{*}}$ or $\highfreq{x_{tar}^{**}}$
\end{tabular}%
}
\vspace{0.4em}
\hrule
\end{minipage}
\end{center}

\section{Main Result Reproduction}

\begin{figure}[!tp]
\centering
\captionof{table}{\small Reproduction and performance comparison of ChordEdit variants on PIE Bench~\cite{ju2023directinversion}. Reproduced values in parentheses report the change relative to the corresponding original result, with \better{green} / \worse{red} indicating \better{better}/\worse{worse}.}
\label{tab:reproduction-comparison}
\vspace{-0.5mm}
{\scriptsize
\setlength{\tabcolsep}{2.5pt}
\renewcommand{\arraystretch}{0.9}
\resizebox{\textwidth}{!}{%
\begin{tabular}{lcccccccccccc}
\toprule
 \multirow{2}{*}{\textbf{Method}} & \multicolumn{3}{c}{\textbf{Timestep}} & \multirow{2}{*}{\makecell{\textbf{Struct.}\\ \textbf{Dist.}$_{10^3}\downarrow$}} & \multicolumn{4}{c}{\textbf{Background Preservation}} & \multicolumn{2}{c}{\textbf{CLIP Semantics}} & \multicolumn{2}{c}{\textbf{Efficiency}} \\
\cmidrule(lr){2-4}\cmidrule(lr){6-9}\cmidrule(lr){10-11}\cmidrule(lr){12-13}
& {\LARGE \lowfreq{$t$}} & {\Large $\delta$} & {\LARGE \highfreq{$t^{**}\ $}} & & \textbf{PSNR}$\uparrow$ & \textbf{MSE}$_{10^3}\downarrow$ & \textbf{SSIM}$_{10^2}\uparrow$ & \textbf{LPIPS}$_{10^3}\downarrow$ & \textbf{Whole}$\uparrow$ & \textbf{Edited}$\uparrow$ & \textbf{Runtime(s)}$\downarrow$ & \textbf{VRAM(MiB)}$\downarrow$ \\
\midrule
\makecell[l]{ChordEdit\\(Naive, SD-Turbo)} & \lowfreq{0.9} & 0.0 & \highfreq{0.3} & 25.44 & 21.38 & 9.73 & 74.39 & 131.30 & 25.11 & 21.96 & 0.38 & 6988 \\
\makecell[l]{ChordEdit\\(Naive w/o prox, SD-Turbo)} & \lowfreq{0.9} & 0.0 & -- & 19.18 & 21.89 & 10.84 & 77.24 & 105.27 & 23.68 & 20.83 & 0.20 & 6988 \\
\textbf{\makecell[l]{ChordEdit\\(Ours, SD-Turbo)}} & \lowfreq{0.9} & 0.15 & \highfreq{0.3} & 16.58 & 22.20 & 6.84 & 75.91 & 128.25 & {25.58} & {22.96} & 0.38 & 6988 \\
\textbf{\makecell[l]{ChordEdit\\(Ours w/o prox, SD-Turbo)}} & \lowfreq{0.9} & 0.15 & -- & 10.37 & 23.89 & 5.05 & 81.24 & 88.36 & 24.97 & 21.87 & 0.20 & 6988 \\
\midrule
\makecell[l]{ChordEdit\\(Naive, SD-Turbo, Reproduced)} & \lowfreq{0.9} & 0.0 & \highfreq{0.3} & \diffcell{45.70}{\worse{(+20.26)}} & \diffcell{20.14}{\worse{(-1.24)}} & \diffcell{13.30}{\worse{(+3.57)}} & \diffcell{73.27}{\worse{(-1.12)}} & \diffcell{155.90}{\worse{(+24.60)}} & \diffcell{25.12}{\better{(+0.01)}} & \diffcell{22.58}{\better{(+0.62)}} & \diffcell{0.51}{\worse{(+0.13)}} & \diffcell{7334}{\worse{(+346)}} \\
\makecell[l]{ChordEdit\\(Naive w/o prox, SD-Turbo, Reproduced)} & \lowfreq{0.9} & 0.0 & -- & \diffcell{45.40}{\worse{(+26.22)}} & \diffcell{20.21}{\worse{(-1.68)}} & \diffcell{13.40}{\worse{(+2.56)}} & \diffcell{75.43}{\worse{(-1.81)}} & \diffcell{143.00}{\worse{(+37.73)}} & \diffcell{24.14}{\better{(+0.46)}} & \diffcell{21.11}{\better{(+0.28)}} & \diffcell{0.51}{\worse{(+0.31)}} & \diffcell{7334}{\worse{(+346)}} \\
\textbf{\makecell[l]{ChordEdit\\(Ours, SD-Turbo, Reproduced)}} & \lowfreq{0.9} & 0.15 & \highfreq{0.3} & \diffcell{29.50}{\worse{(+12.92)}} & \diffcell{22.64}{\better{(+0.44)}} & \diffcell{8.00}{\worse{(+1.16)}} & \diffcell{76.75}{\better{(+0.84)}} & \diffcell{118.50}{\better{(-9.75)}} & \diffcell{24.82}{\worse{(-0.76)}} & \diffcell{22.16}{\worse{(-0.80)}} & \diffcell{0.51}{\worse{(+0.13)}} & \diffcell{7334}{\worse{(+346)}} \\
\textbf{\makecell[l]{ChordEdit\\(Ours w/o prox, SD-Turbo, Reproduced)}} & \lowfreq{0.9} & 0.15 & -- & \diffcell{21.90}{\worse{(+11.53)}} & \diffcell{23.90}{\better{(+0.01)}} & \diffcell{6.40}{\worse{(+1.35)}} & \diffcell{81.35}{\better{(+0.11)}} & \diffcell{85.00}{\better{(-3.36)}} & \diffcell{23.68}{\worse{(-1.29)}} & \diffcell{20.84}{\worse{(-1.03)}} & \diffcell{0.51}{\worse{(+0.31)}} & \diffcell{7334}{\worse{(+346)}} \\
\bottomrule
\end{tabular}%
}
}

\vspace{1mm}
\captionof{table}{\small Ablation of chord transport field and proximal refinement. The first block reports the original numbers, and the second block reports our reproduced numbers. $\Delta$ rows compute the effect of adding proximal refinement (w/ prox $-$ w/o prox), while $\Delta_\delta$ columns compute the effect of the chord transport field by comparing Ours ($\delta=0.15$) against Naive ($\delta=0$).}
\label{tab:ablation-field-refinement}
\vspace{0.5mm}
{\scriptsize
\setlength{\tabcolsep}{4.5pt}
\renewcommand{\arraystretch}{0.9}
\resizebox{\textwidth}{!}{%
\begin{tabular}{lcccccccccc}
\toprule
\multirow{2}{*}{\textbf{Method}} & \multicolumn{3}{c}{\textbf{Timestep}} & \multicolumn{2}{c}{\textbf{Naive ($\delta=0$)}} & \multicolumn{2}{c}{\textbf{Ours ($\delta=0.15$)}} & \multicolumn{2}{c}{\textbf{$\Delta_\delta$ ($0.15-0$)}} & \textbf{NFE} \\
\cmidrule(lr){2-4}\cmidrule(lr){5-6}\cmidrule(lr){7-8}\cmidrule(lr){9-10}
 &  {\LARGE \lowfreq{$t$}} & {\Large $\delta$} & {\LARGE \highfreq{$t^{**}\ $}} & \textbf{PSNR}$\uparrow$ & \textbf{CLIP-Edited}$\uparrow$ & \textbf{PSNR}$\uparrow$ & \textbf{CLIP-Edited}$\uparrow$ & \textbf{PSNR}$\uparrow$ & \textbf{CLIP-Edited}$\uparrow$ & \\
\midrule
w/o prox $\lowfreq{x_{tar}^*}$ & \lowfreq{0.9} & 0/0.15 & -- & 21.89 & 20.83 & \textbf{23.89} & 21.87 & \better{$+2.00$} & \better{$+1.04$} & 1 \\
w/ prox $\highfreq{x_{tar}^{**}}$ & \lowfreq{0.9} & 0/0.15 & \highfreq{0.3} & 21.38 & 21.96 & 22.20 & \textbf{22.96} & \better{$+0.82$} & \better{$+1.00$} & 2 \\
$\Delta$ (w/ prox $-$ w/o prox) & \lowfreq{0.9} & 0/0.15 & --$\to$\highfreq{0.3} & \worse{$-0.51$} & \better{$+1.13$} & \worse{$-1.69$} & \better{$+1.09$} & \worse{$-1.18$} & \worse{$-0.04$} & $+1$ \\

\midrule
w/o prox $\lowfreq{x_{tar}^*}$ (reproduced) & \lowfreq{0.9} & 0/0.15 & -- & 20.21 & 21.11 & \textbf{23.90} & 20.84 & \better{$+3.69$} & \worse{$-0.27$} & 1 \\
w/ prox $\highfreq{x_{tar}^{**}}$ (reproduced) & \lowfreq{0.9} & 0/0.15 & \highfreq{0.3} & 20.14 & \textbf{22.58} & 22.64 & 22.16 & \better{$+2.50$} & \worse{$-0.42$} & 2 \\
$\Delta$ (w/ prox $-$ w/o prox, reproduced) & \lowfreq{0.9} & 0/0.15 & --$\to$\highfreq{0.3} & \worse{$-0.07$} & \better{$+1.47$} & \worse{$-1.26$} & \better{$+1.32$} & \worse{$-1.19$} & \worse{$-0.15$} & $+1$ \\

\bottomrule
\end{tabular}%
}
}


\vspace{5mm}
\includegraphics[width=0.85\textwidth]{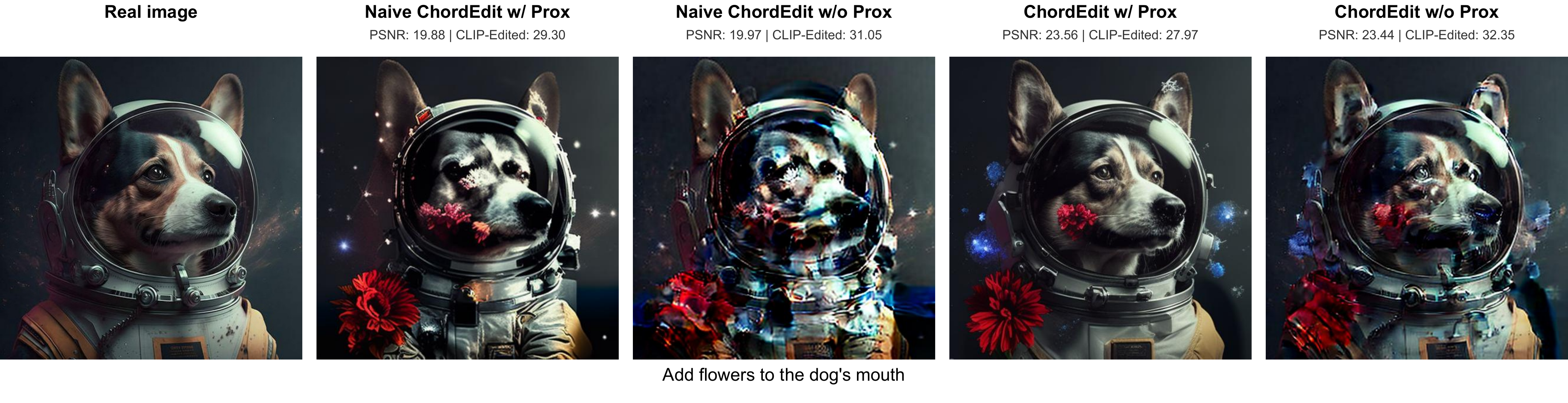}
\vspace{-1mm}
\includegraphics[width=0.85\textwidth]{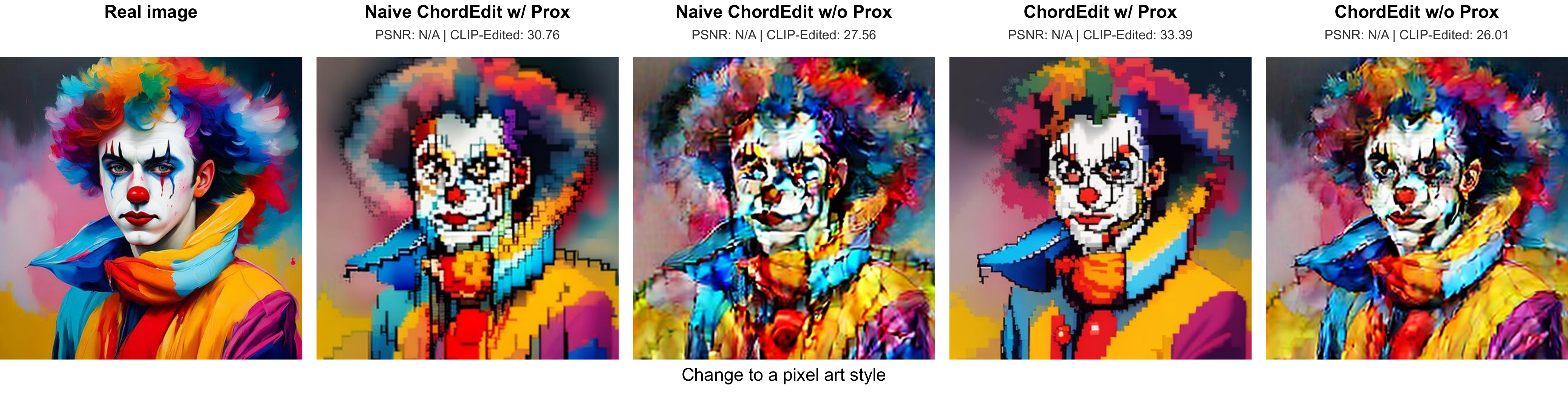}
\vspace{-1mm}
\includegraphics[width=0.85\textwidth]{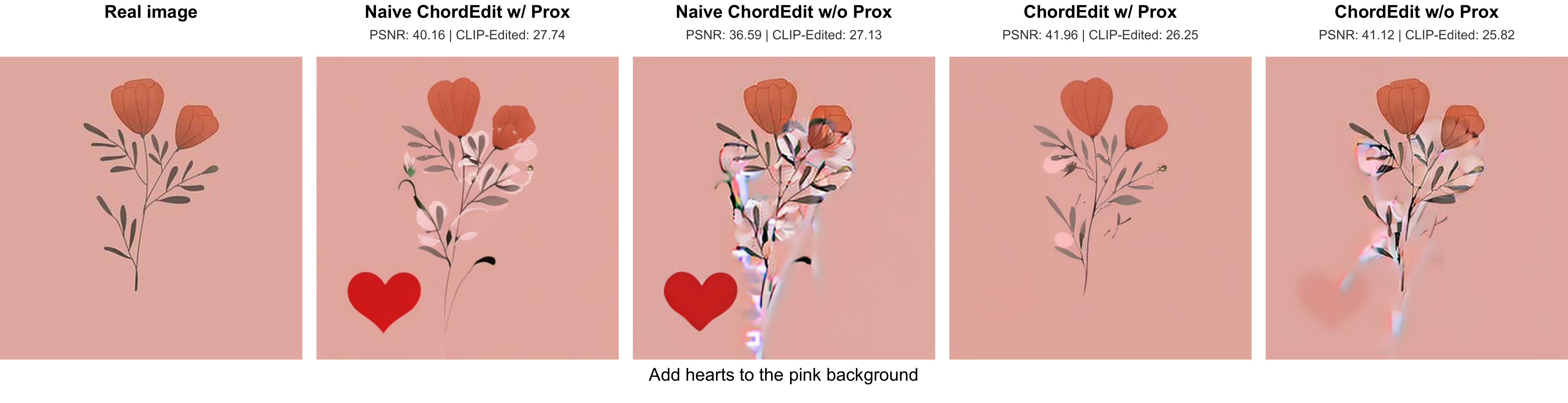}
\vspace{-3mm}
\caption{\small Qualitative examples comparing ChordEdit variants. Each row shows one source example with the corresponding ablation columns, illustrating how the choice of timestep and proximal refinement changes the edited result.}
\label{fig:naive-chordedit-qualitative}
\end{figure}

\textbf{Experimental setting.} We reproduce ChordEdit on PIE Bench~\cite{ju2023directinversion} using the authors' released official code, evaluation scripts, and default settings. All experiments are conducted on a single NVIDIA RTX A6000 GPU. We report both the original numbers and our reproduced results under the same configuration, including the one-step setting without proximal refinement and the two-step setting with proximal refinement.

\noindent\fcolorbox{black}{gray!10}{%
\begin{minipage}{\dimexpr\textwidth-2\fboxsep-2\fboxrule\relax}
\textbf{Finding 1: From Table~\ref{tab:reproduction-comparison}, naive reproduction differs from the original results, especially on semantic metrics such as CLIP-Whole and CLIP-Edited.}
\end{minipage}%
}

For the naive baselines, Table~\ref{tab:reproduction-comparison} compares the original results in rows 1--2 with our reproduced results in rows 5--6. Our reproduced results are worse in background preservation than the original numbers, as reflected by lower PSNR and higher LPIPS. However, semantic alignment improves, especially on the edited region: CLIP-Edited increases from $21.96$ to $22.58$ for the naive variant with proximal refinement, and from $20.83$ to $21.11$ for the naive variant without proximal refinement.

For ChordEdit, however, \textbf{\textit{our reproduced results show a clear gap in semantic metrics regardless of whether proximal refinement is used}}. With proximal refinement, CLIP-Whole drops from $25.58$ to $24.82$ ($-0.76$), and CLIP-Edited drops from $22.96$ to $22.16$ ($-0.80$). Without proximal refinement, the gap is even larger: CLIP-Whole drops from $24.97$ to $23.68$ ($-1.29$), and CLIP-Edited drops from $21.87$ to $20.84$ ($-1.03$). These consistent decreases suggest that, although our reproduction partially preserves background metrics for the ChordEdit rows, it does not fully recover the semantic alignment reported in the original results.

\noindent\fcolorbox{black}{cyan!10}{%
\begin{minipage}{\dimexpr\textwidth-2\fboxsep-2\fboxrule\relax}
\textbf{Finding 2: From Table~\ref{tab:ablation-field-refinement}, our reproduction and the original paper lead to different conclusions about the semantic effect of chord transport: the original paper shows semantic improvement, while our reproduction shows a semantic decline.}
\end{minipage}%
}

Table~\ref{tab:ablation-field-refinement} summarizes the key ablation from Table~\ref{tab:reproduction-comparison}. From the last three columns, the effect of the chord transport field is less consistent. In the original results, changing from Naive ($\delta=0$) to Ours ($\delta=0.15$) improves both background preservation and semantic alignment: without proximal refinement, PSNR increases by $2.00$ and \better{\textit{CLIP-Edited increases by $1.04$}}; with proximal refinement, PSNR increases by $0.82$ and \better{\textit{CLIP-Edited increases by $1.00$}}. However, in our reproduction, the chord transport field improves preservation but reduces semantic alignment: PSNR increases by $3.69$ without proximal refinement and by $2.50$ with proximal refinement, while \worse{\textit{CLIP-Edited decreases by $0.27$ and $0.42$}}, respectively.

\noindent\fcolorbox{black}{violet!10}{%
\begin{minipage}{\dimexpr\textwidth-2\fboxsep-2\fboxrule\relax}
\textbf{Finding 3: From Table~\ref{tab:ablation-field-refinement}, on proximal refinement, our reproduction is consistent with the original paper: it consistently improves semantics with minimal background-preservation loss.}
\end{minipage}%
}

From the third and sixth rows of Table~\ref{tab:ablation-field-refinement}, the original and reproduced results give a consistent conclusion about proximal refinement: it slightly reduces background preservation but substantially improves edited-area semantic alignment. In the original results, adding proximal refinement \worse{\textit{decreases PSNR by $0.51$ for the naive setting and by $1.69$ for ChordEdit}}, while \better{\textit{increasing CLIP-Edited by $1.13$ and $1.09$}}, respectively. The reproduced results show the same pattern: \worse{\textit{PSNR decreases by $0.07$ and $1.26$}}, while \better{\textit{CLIP-Edited increases by $1.47$ and $1.32$}}.

\noindent\fcolorbox{black}{gray!10}{%
\begin{minipage}{\dimexpr\textwidth-2\fboxsep-2\fboxrule\relax}
\textbf{Reproducibility note.} The reproduced numbers above may vary with implementation details, random seeds, library versions, and hardware differences such as the GPU type. We welcome the community to rerun the evaluation, report discrepancies, and help refine or correct our reproduced results.
\end{minipage}%
}

\begin{figure}[p]
\centering
\includegraphics[width=0.91\textwidth]{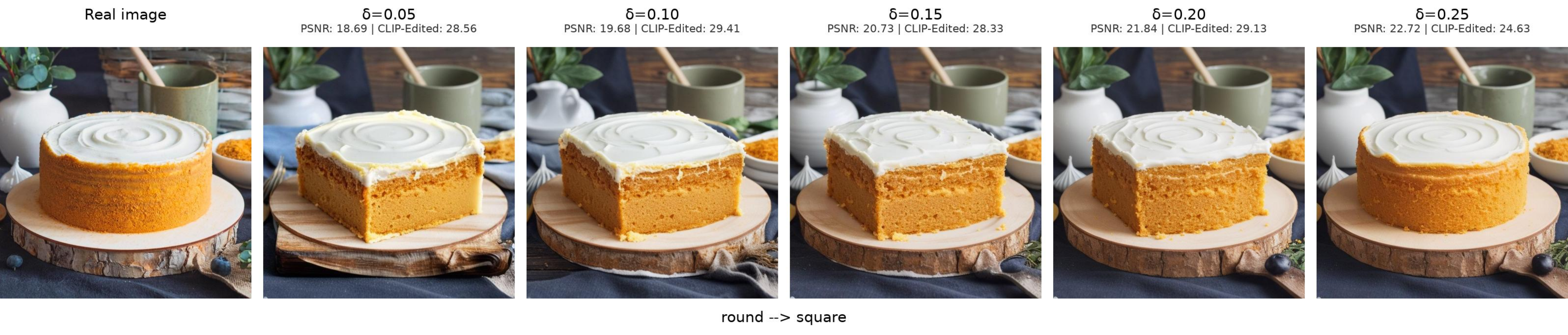}
\vspace{0.5mm}
\includegraphics[width=0.91\textwidth]{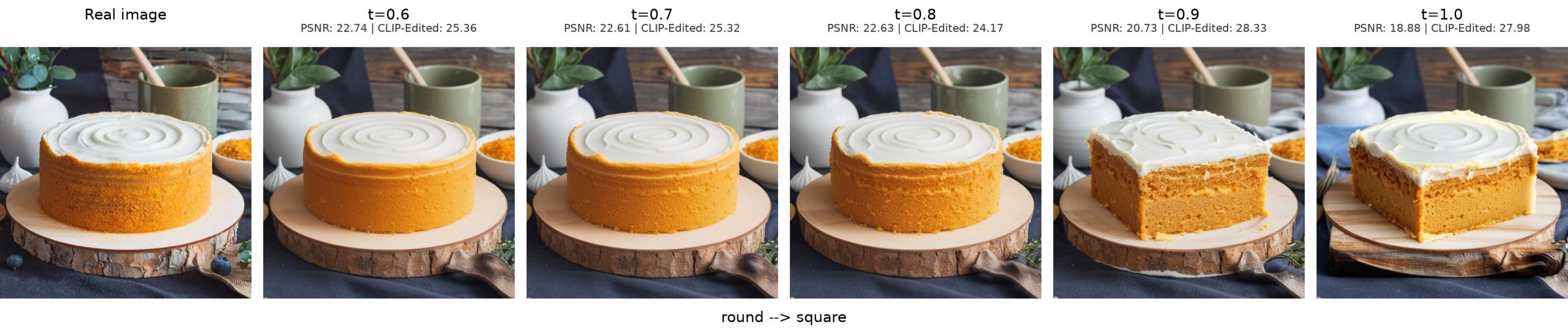}
\vspace{0.5mm}
\includegraphics[width=0.91\textwidth]{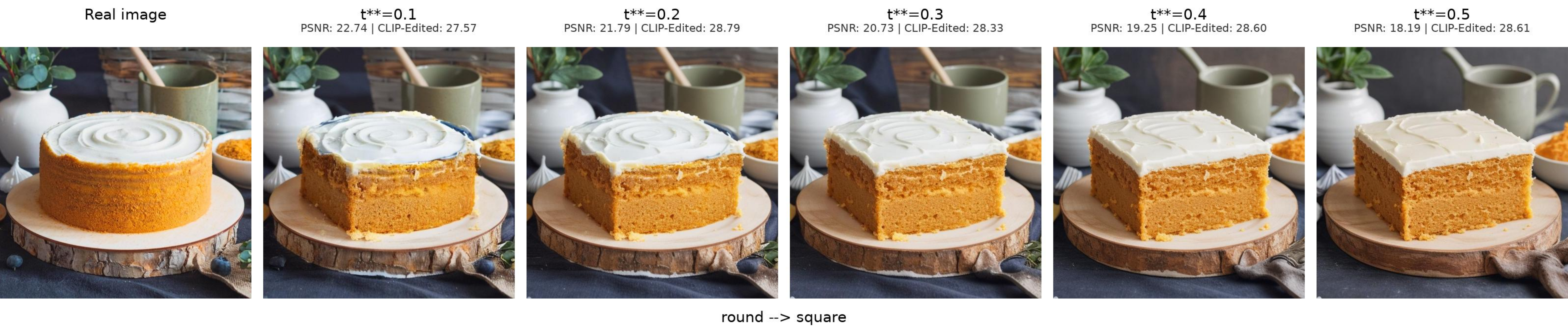}
\vspace{-3mm}
\caption{\small Anchor example 000000000001: grouped qualitative sweeps for the chord window $\delta$, editing timestep $t$, and proximal-refinement timestep $t^{**}$. Grouping the three sweeps for the same source image makes it easier to compare how each parameter changes the edited result.}
\label{fig:anchor-000000000001-grouped-sweeps}
\end{figure}

\begin{figure}[p]
\centering
\includegraphics[width=0.91\textwidth]{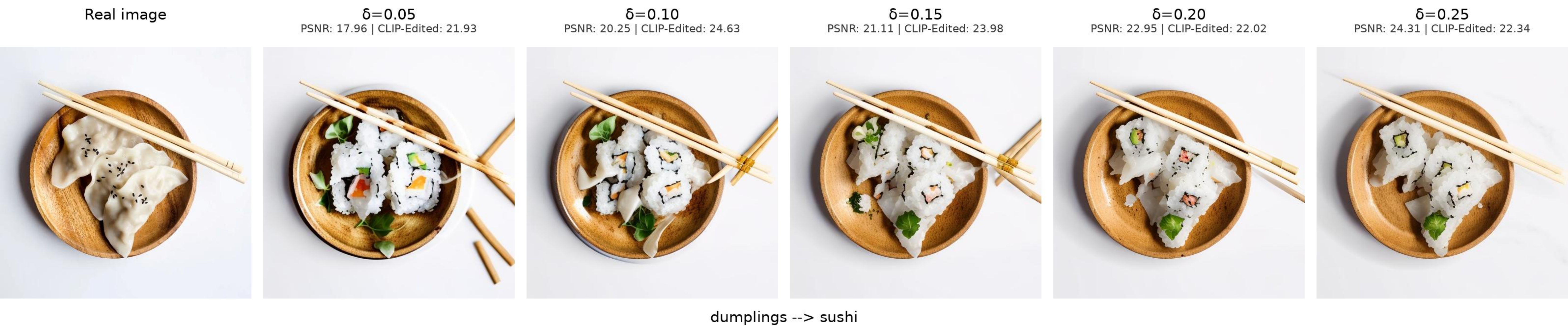}
\vspace{0.5mm}
\includegraphics[width=0.91\textwidth]{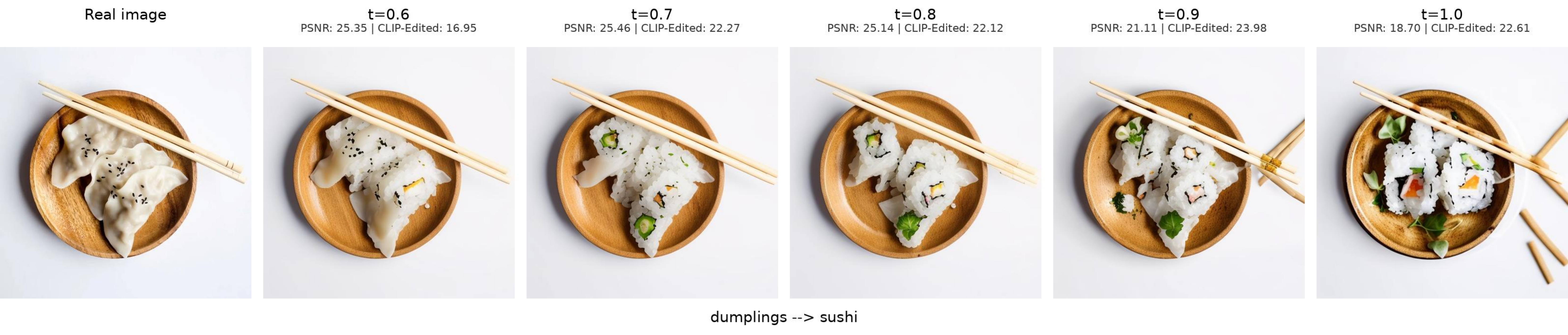}
\vspace{0.5mm}
\includegraphics[width=0.91\textwidth]{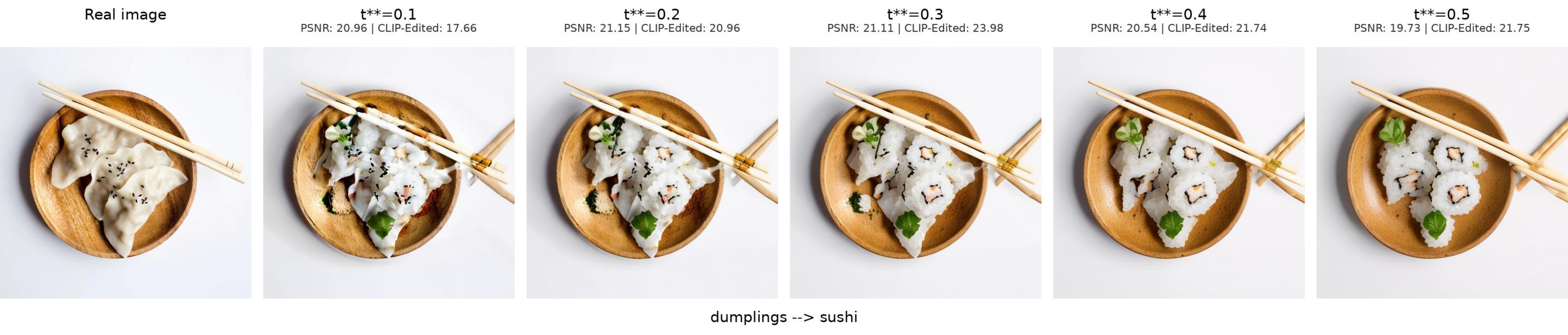}
\vspace{-3mm}
\caption{\small Anchor example 000000000013: grouped qualitative sweeps for the chord window $\delta$, editing timestep $t$, and proximal-refinement timestep $t^{**}$. Grouping the three sweeps for the same source image makes it easier to compare how each parameter changes the edited result.}
\label{fig:anchor-000000000013-grouped-sweeps}
\end{figure}



\newpage
\section{Ablation Study Reproduction}

These findings motivate a broader question: in one-step or few-step image editing, what component is actually doing the work---the chord field itself $\delta$, the selected timestep \lowfreq{$t$}, the proximal refinement step \highfreq{$t^{**}$}, or their interaction? 

\begin{figure}[h]
\centering
\includegraphics[width=0.96\textwidth]{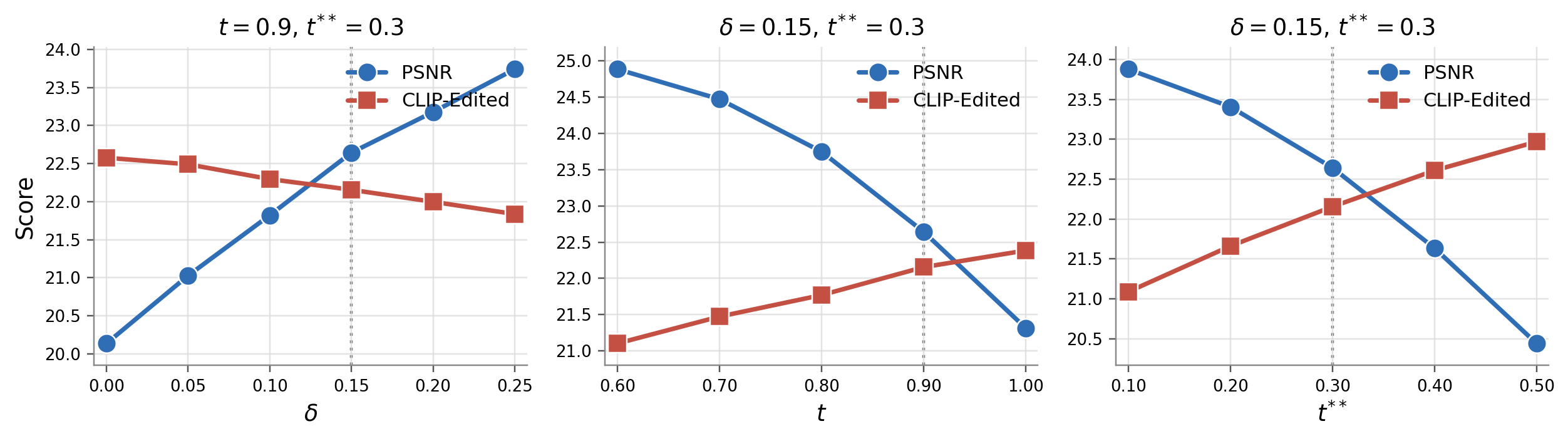}
\vspace{-4mm}
\caption{\small Single-parameter ablation study for one/few-step image editing. Each plot varies one parameter while keeping the other two fixed at the default values $\delta=0.15$, \lowfreq{$t=0.9$}, and \highfreq{$t^{**}=0.3$}, and reports the resulting PSNR and CLIP-Edited scores. This isolates the effect of the chord window $\delta$, the editing timestep $t$, and the proximal refinement timestep $t^{**}$ on fidelity and semantic alignment. }
\label{fig:single-parameter-ablation}
\end{figure}

\textbf{Single-parameter ablation studies (Figure~\ref{fig:single-parameter-ablation}).} To examine this more rigorously, we conduct single-parameter ablation studies: we fix two variables and vary only the remaining one. For example, when studying the chord window $\delta$, we keep the other settings fixed (\lowfreq{$t=0.9$}, \highfreq{$t^{**}=0.3$}) and evaluate how PSNR and CLIP-Edited change as $\delta$ varies. The resulting fidelity--semantics curves are shown in the leftmost subfigure of Figure~\ref{fig:single-parameter-ablation}. 

Figure~\ref{fig:single-parameter-ablation} shows three main trends. First, varying the chord window $\delta$ has a non-monotonic effect: the optimal value is $\delta=0.15$, while moving either smaller or larger weakens the fidelity--semantics trade-off. Second, varying the editing timestep \lowfreq{$t$} reveals that the optimal value is \lowfreq{$t=0.9$}, indicating that the first editing step is highly sensitive to the chosen noise level. Third, increasing the proximal refinement time \highfreq{$t^{**}$} steadily improves CLIP-Edited but lowers PSNR, showing that the second target-conditioned step strengthens semantic alignment at the cost of background preservation. The optimal value is \highfreq{$t^{**} = 0.3$}.

\textbf{Grouped qualitative visualizations (Figures~\ref{fig:anchor-000000000001-grouped-sweeps} - \ref{fig:anchor-000000000013-grouped-sweeps}).} We further provide image-wise visualizations for the same examples, where each block compares the effects of varying $\delta$, \lowfreq{$t$}, and \highfreq{$t^{**}$} on the same source image. These visualizations make the trends in Figure~\ref{fig:single-parameter-ablation} more concrete. For the $\delta$ sweep, smaller $\delta$ corresponds to a larger effective noise level and moves the result closer to a newly target-guided generation, while larger $\delta$ lowers the effective noise level and keeps the output closer to the source image. For the editing timestep sweep, smaller \lowfreq{$t$} means lower noise and stronger source preservation, whereas larger \lowfreq{$t$} means higher noise and a stronger target-guided edit. A similar trend appears for \highfreq{$t^{**}$}: increasing the proximal-refinement timestep injects more target-conditioned generation and improves semantic strength, but also makes the result deviate more from the source.

\noindent\fcolorbox{black}{gray!10}{%
\begin{minipage}{\dimexpr\textwidth-2\fboxsep-2\fboxrule\relax}
\textbf{Finding 4: Figures~\ref{fig:anchor-000000000001-grouped-sweeps} - \ref{fig:single-parameter-ablation} confirm that the default ChordEdit setting is an optimal parameter choice in our reproduction. This raises a more fundamental question: what is the chord window actually doing?}
\end{minipage}%
}

We return to this question in the next section and examine whether the benefit comes from the chord window itself, or from the effective timestep implicitly selected by this window.

\begin{figure}[!tbp]
\centering
\captionof{table}{\small Simplification test for the chord control field. The default reproduced ChordEdit setting uses \lowfreq{$t=0.9$} and $\delta=0.15$, while the simplified setting directly uses the dominant term suggested by Eq.~\ref{eq:chord-control-field-rf}, i.e., \lowfreq{$t=0.75$} and $\delta=0$.}
\label{tab:simplified-chord-field}
\vspace{-0.5mm}
{\large
\setlength{\tabcolsep}{4pt}
\renewcommand{\arraystretch}{1.1}
\resizebox{\textwidth}{!}{%
\begin{tabular}{lcccccccccccc}
\toprule
 {\textbf{Method}} & \multicolumn{3}{c}{\textbf{Timestep}} & \multirow{2}{*}{\makecell{\textbf{Struct.}\\ \textbf{Dist.}$_{10^3}\downarrow$}} & \multicolumn{4}{c}{\textbf{Background Preservation}} & \multicolumn{2}{c}{\textbf{CLIP Semantics}} & \multicolumn{2}{c}{\textbf{Efficiency}} \\
\cmidrule(lr){2-4}\cmidrule(lr){6-9}\cmidrule(lr){10-11}\cmidrule(lr){12-13}
& {\LARGE \lowfreq{$t$}} & {\Large $\delta$} & {\LARGE \highfreq{$t^{**}\ $}} & & \textbf{PSNR}$\uparrow$ & \textbf{MSE}$_{10^3}\downarrow$ & \textbf{SSIM}$_{10^2}\uparrow$ & \textbf{LPIPS}$_{10^3}\downarrow$ & \textbf{Whole}$\uparrow$ & \textbf{Edited}$\uparrow$ & \textbf{Runtime(s)}$\downarrow$ & \textbf{VRAM(MiB)}$\downarrow$ \\
\midrule
\makecell[l]{ChordEdit\\(default, reproduced)} & \lowfreq{0.9} & 0.15 & \highfreq{0.3} & 29.50 & \textbf{22.64} & \textbf{8.00} & \textbf{76.75} & \textbf{118.50} & 24.82 & 22.16 & 0.51 & 7334 \\
\makecell[l]{ChordEdit\\(naive, reproduced)} & \lowfreq{0.75} & 0.0 & \highfreq{0.3} & 31.40 & 22.36 & 8.60 & 76.30 & 123.00 & \textbf{25.43} & \textbf{22.23} & - & - \\
\bottomrule
\end{tabular}%
}
}

\vspace{3mm}
\includegraphics[width=0.48\textwidth]{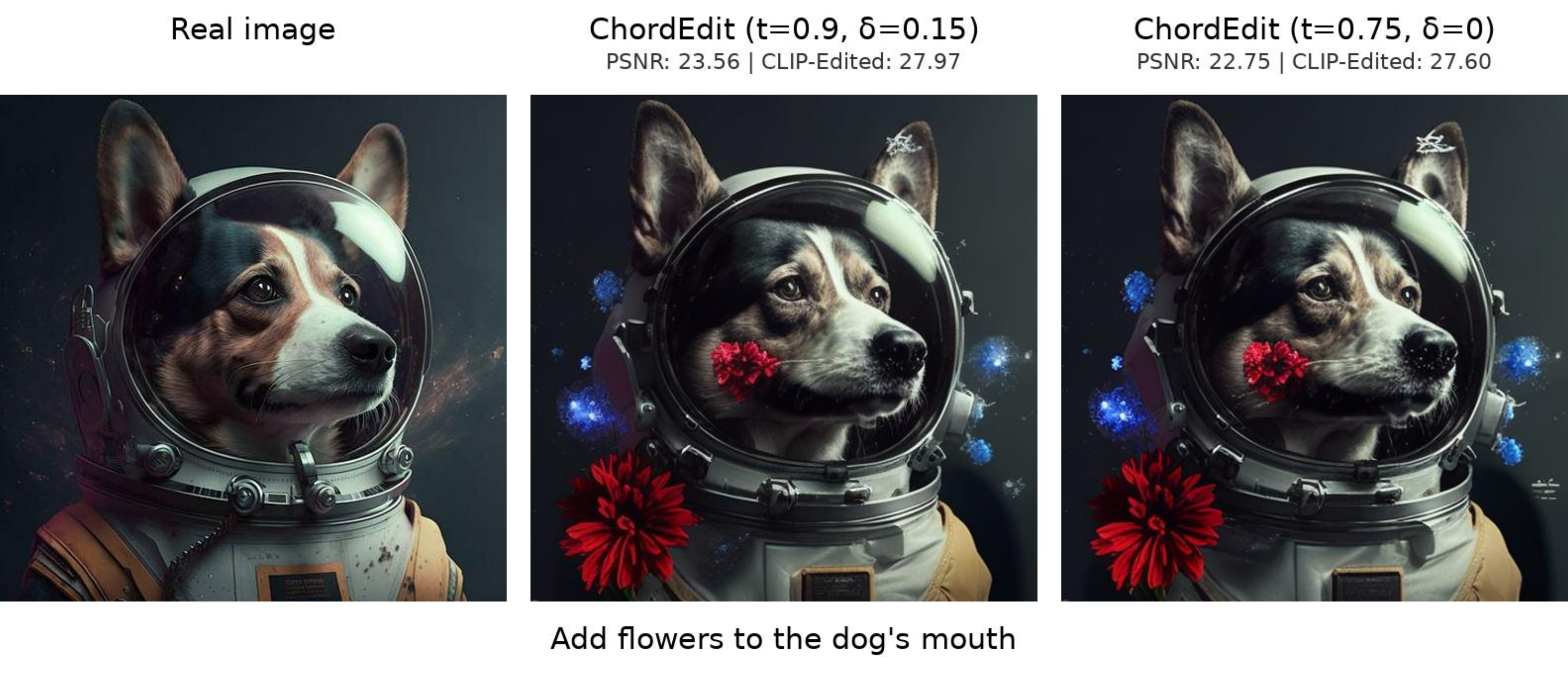}
\vspace{1mm}
\includegraphics[width=0.48\textwidth]{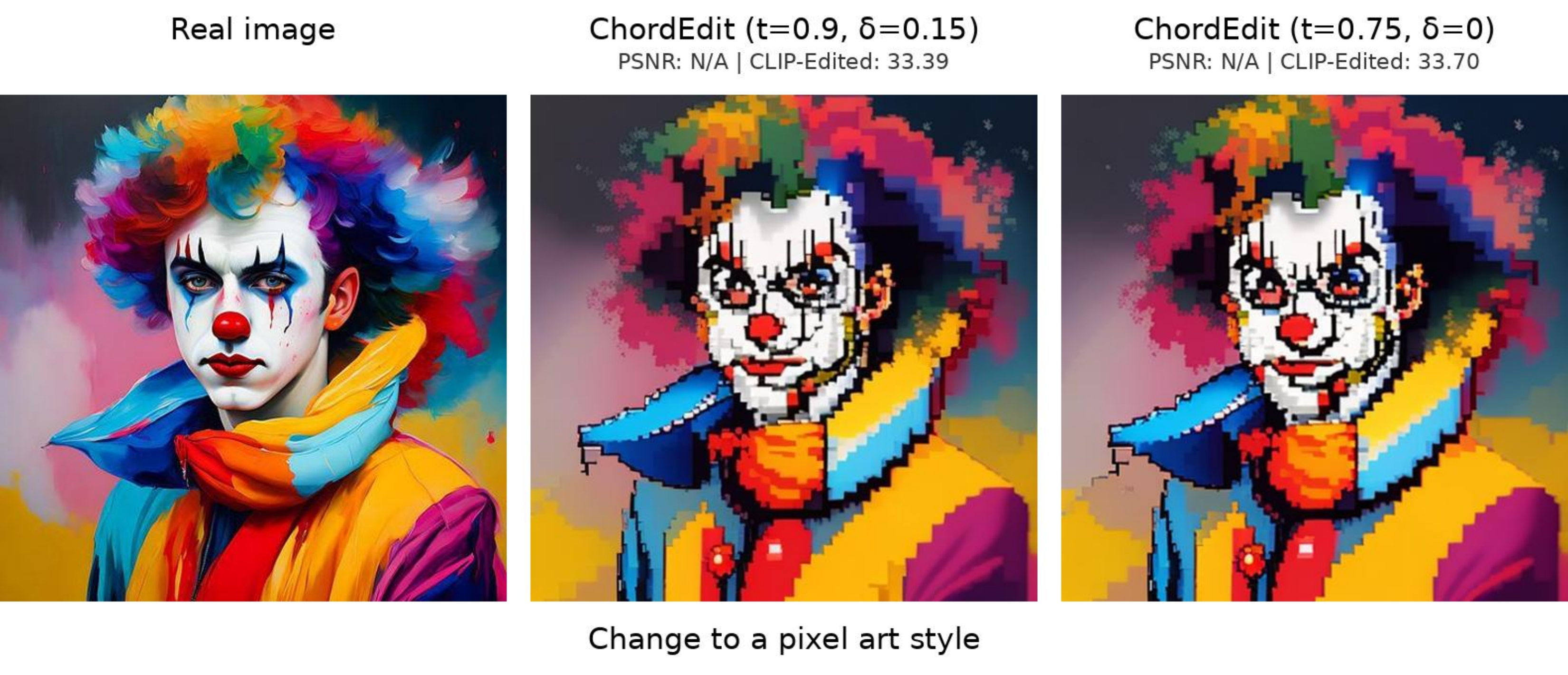}
\vspace{1mm}
\includegraphics[width=0.48\textwidth]{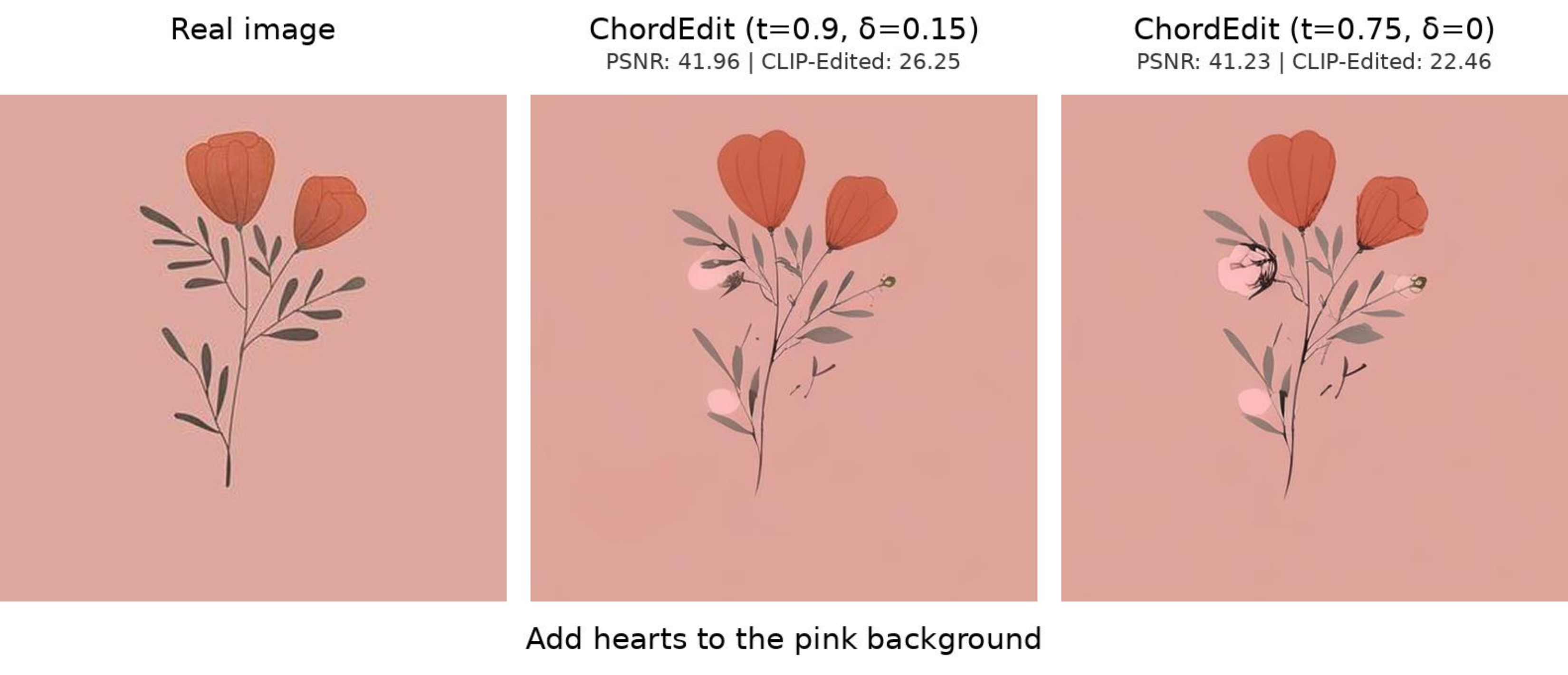}
\vspace{1mm}
\includegraphics[width=0.48\textwidth]{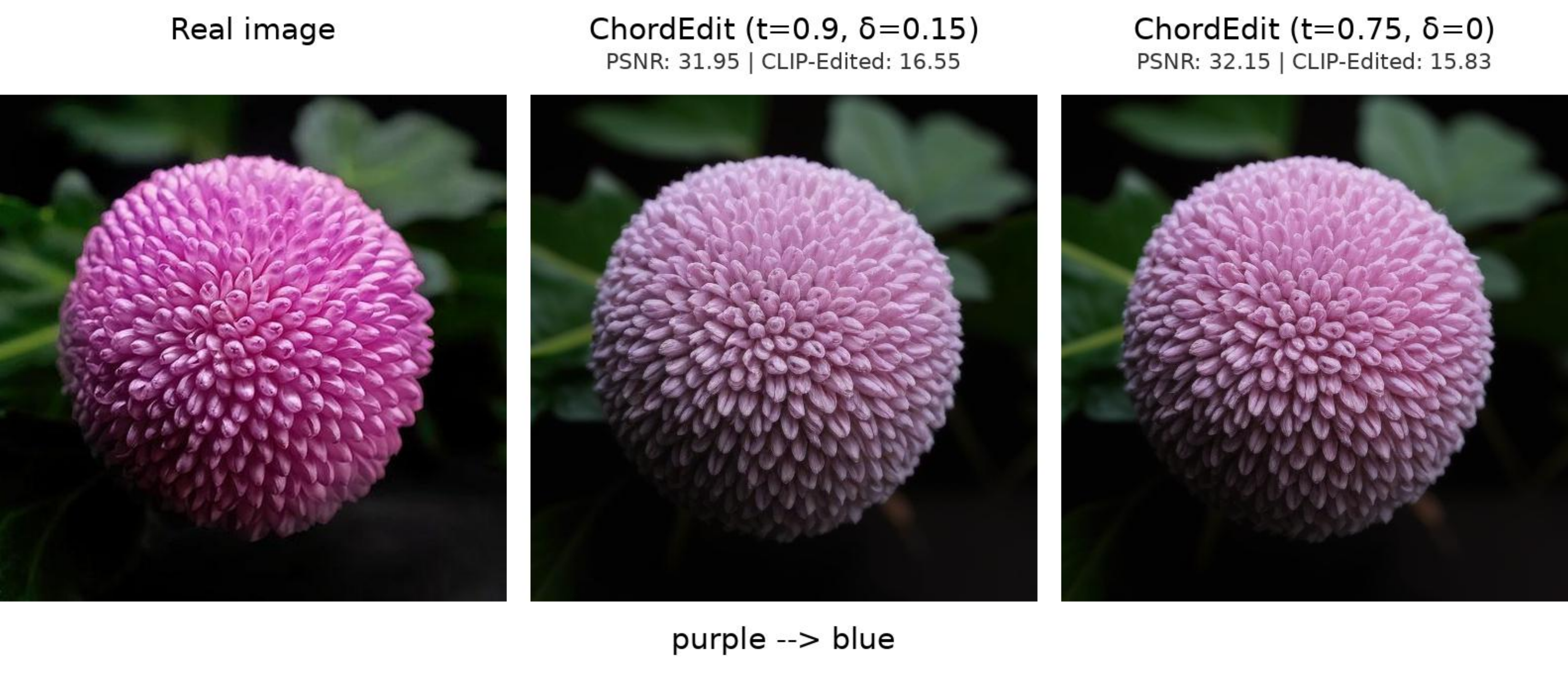}
\vspace{1mm}
\includegraphics[width=0.48\textwidth]{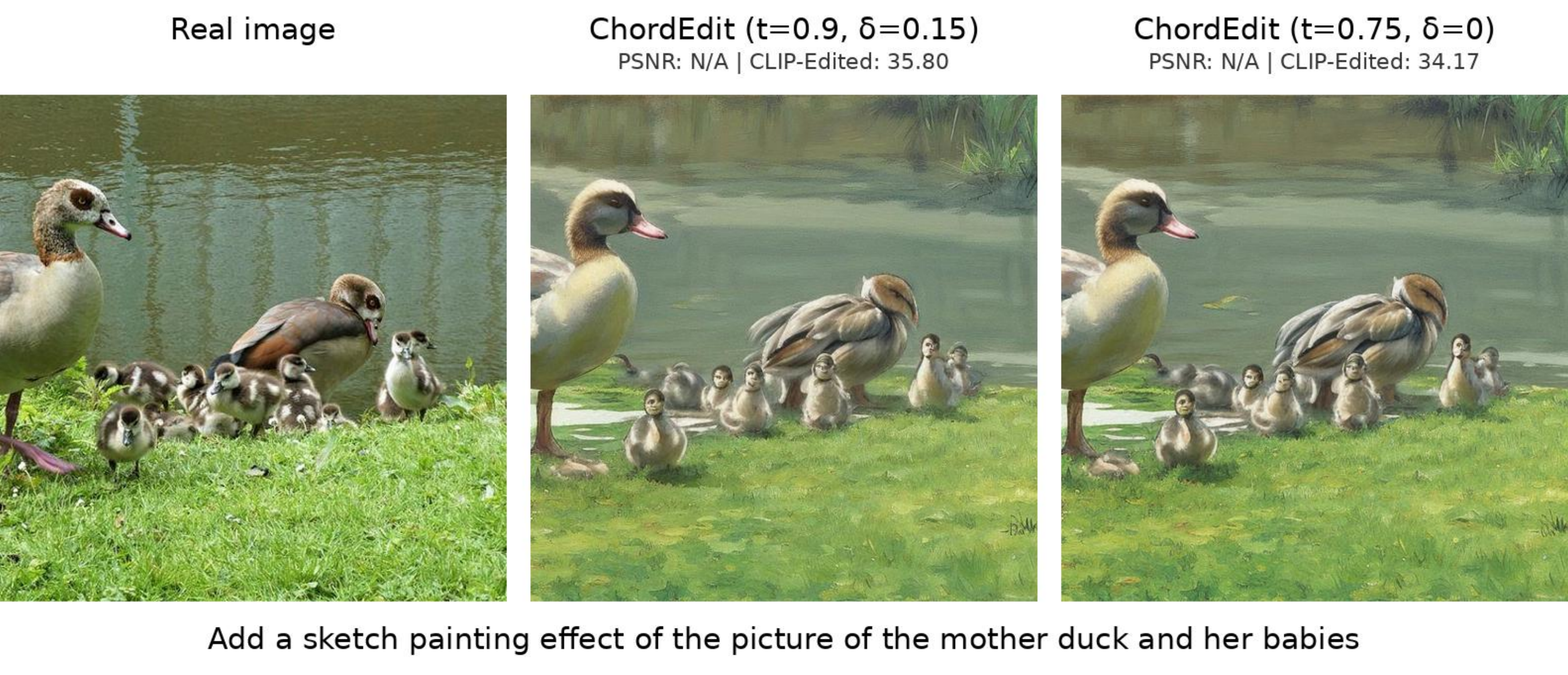}
\includegraphics[width=0.48\textwidth]{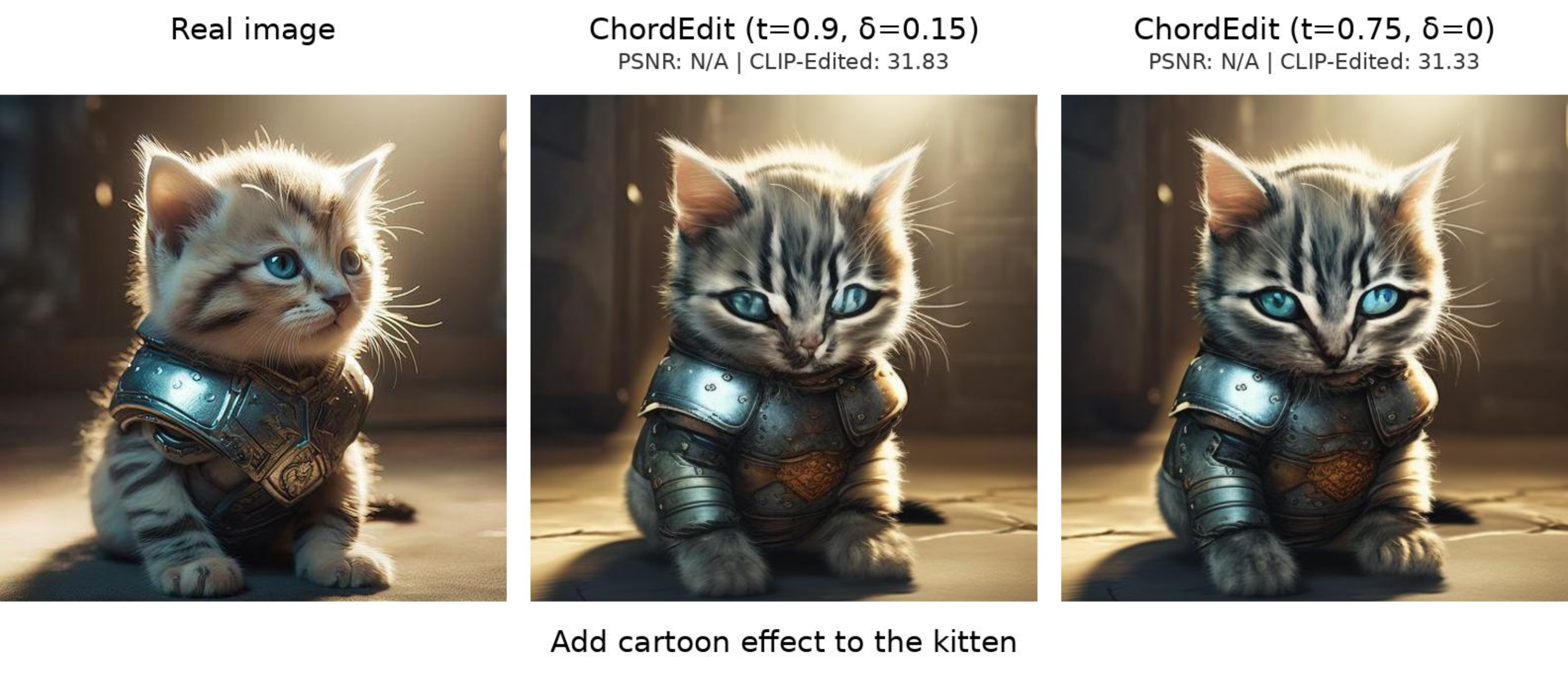}
\vspace{0mm}
\caption{\small Qualitative comparison between the default ChordEdit setting ($t=0.9,\delta=0.15$) and the simplified dominant-velocity setting ($t=0.75,\delta=0$). The two settings produce nearly identical visual results, supporting the interpretation that the default chord update is dominated by the velocity term at the effective timestep $t-\delta=0.75$.}
\label{fig:default-vs-t075-qualitative}

\vspace{2mm}
\includegraphics[width=0.45\textwidth]{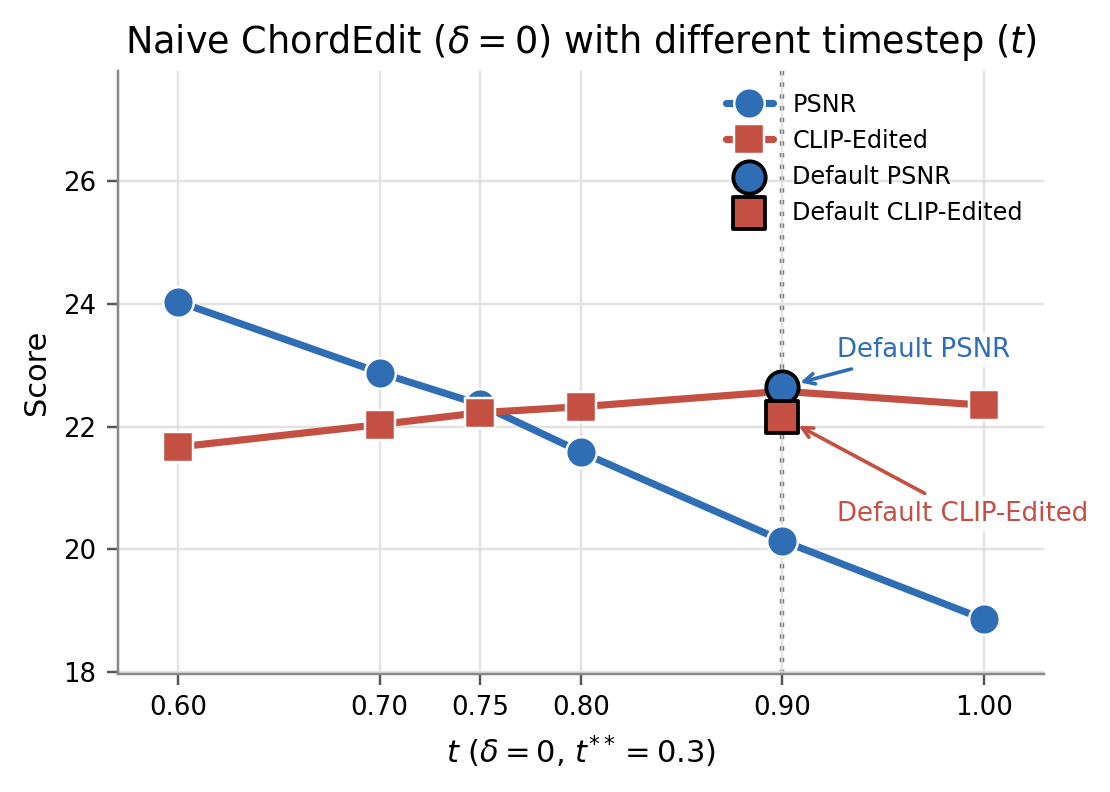}
\vspace{-1mm}
\caption{\small Ablation of the naive ChordEdit with $\delta=0$ under proximal refinement (\highfreq{$t^{**}=0.3$}). Varying the starting timestep shows that \lowfreq{$t=0.75$} gives the best fidelity--semantics trade-off, matching the dominant effective timestep induced by the default ChordEdit setting.}
\label{fig:simplification-ablation-delta0}

\end{figure}
\newpage
\section{Simplification}

Based on Eq.~(\ref{eq:chord-control-field-rf}) and the default ChordEdit setting used in Table~\ref{tab:reproduction-comparison}, we substitute \lowfreq{$t=0.9$} and $\delta=0.15$. Since \lowfreq{$t-\delta=0.75$}, the chord control field becomes
\begin{equation}
    \hat{u}_{0.9}(x_{0.9})
    \ = \frac{0.9\Delta\mathbf{v}(x_{0.75},0.75)
    + 0.15\Delta\mathbf{v}(x_{0.9},0.9)}{1.05}
    \ = \frac{6}{7}\Delta\mathbf{v}(x_{0.75},0.75)
    + \frac{1}{7}\Delta\mathbf{v}(x_{0.9},0.9).
\end{equation}
Thus, under the default setting, the update direction is dominated by the velocity difference at the shifted timestep \lowfreq{$t-\delta=0.75$}, while the current-state velocity difference at \lowfreq{$t=0.9$} contributes only 1/7 of the final chord control field. In this sense, the chord window mainly acts as a timestep shift from \lowfreq{$t$} to \lowfreq{$t-\delta$}, rather than as a fully independent transport path.

Based on this observation, we compare the default setting $(\lowfreq{t=0.9},\delta=0.15, \highfreq{t^{**}=0.3)}$ with a simplified setting that removes the chord window and directly uses the dominant timestep $(\lowfreq{t=0.75},\delta=0, \highfreq{t^{**}=0.3})$. Table~\ref{tab:simplified-chord-field} reports the resulting performance. Surprisingly, this simplified setting achieves performance very close to the default reproduced ChordEdit setting. It slightly reduces background preservation metrics, e.g., PSNR changes from $22.64$ to $22.36$ and LPIPS changes from $118.50$ to $123.00$, but it slightly improves semantic metrics, with CLIP-Whole increasing from $24.82$ to $25.43$ and CLIP-Edited increasing from $22.16$ to $22.23$.

Beyond this two-point comparison, we further conduct an ablation study with $\delta=0$ and different starting timesteps to understand the fidelity--semantics trend. As shown in Figure~\ref{fig:simplification-ablation-delta0}, when proximal refinement is used (left subfigure, \highfreq{$t^{**}=0.3$}), the best trade-off appears around \lowfreq{$t=0.75$}, which aligns with the dominant timestep derived from the default ChordEdit update above. 
%
{Together, these findings suggest that the empirical gains may not come solely from the full chord transport formulation. Instead, choosing a suitable noise level appears to be a key factor, together with whether a second target-conditioned refinement step is used.}

\noindent\fcolorbox{black}{cyan!10}{%
\begin{minipage}{\dimexpr\textwidth-2\fboxsep-2\fboxrule\relax}
\textbf{Finding 5: The chord window mainly shifts the effective editing timestep from $t$ to $t-\delta$. This reveals that the key factor is not the chord path itself, but choosing a good effective timestep.}
\end{minipage}%
}

\section{New Insights}

ChordEdit is a highly insightful work because it exposes a useful way to think about fast image editing: instead of treating one-step editing as a black-box jump from source to target, it frames the update as transport along a noise-conditioned path. Our simplification does not diminish this contribution. Rather, by rewriting the mechanism in more intuitive terms, we aim to make the core idea easier to understand, analyze, and reuse. We believe such a simplified interpretation is also valuable for community development, because it separates the empirical recipe from the underlying design principle and suggests how future methods might combine different noise levels more deliberately.

\noindent\fcolorbox{black}{cyan!10}{%
\begin{minipage}{\dimexpr\textwidth-2\fboxsep-2\fboxrule\relax}
\textbf{Insight 1: High-noise editing controls low-frequency transport.}
\end{minipage}%
}

The first stage should be understood as a coarse semantic transport step. At a relatively high noise level \lowfreq{$t$}, the model state contains less reliable spatial detail and more global ambiguity, so the conditional velocity difference $\Delta\mathbf{v}(x_t,t)=\mathbf{v}(x_t,t,c_{tar})-\mathbf{v}(x_t,t,c_{src})$ mainly changes the low-frequency components of the image: object identity, layout, pose, and other large-scale semantic structure. This explains why choosing the effective timestep is crucial. If the noise level is too low, the edit cannot overcome the source image; if it is too high, the result becomes closer to target-guided regeneration. A good first-stage timestep therefore acts as a controlled low-frequency bridge from the source prompt to the target prompt.

\noindent\fcolorbox{black}{violet!10}{%
\begin{minipage}{\dimexpr\textwidth-2\fboxsep-2\fboxrule\relax}
\textbf{Insight 2: Low-noise refinement restores high-frequency target details.}
\end{minipage}%
}

The second stage plays a complementary role. After the coarse edit has already moved the image toward the target semantics, proximal refinement uses a smaller noise level \highfreq{$t^{**}$} and a purely target-conditioned velocity field to add the high-frequency evidence that the first stage cannot reliably specify. This step sharpens target-specific appearance, texture, local details, and fine semantic cues, thereby pulling the edited image closer to the target prompt without completely resampling the global structure. In this view, ChordEdit-like editing is not a single uniform transformation, but a two-level process: high-noise dynamics perform low-frequency semantic transport, while low-noise dynamics refine high-frequency target alignment.

\noindent\fcolorbox{black}{yellow!10}{%
\begin{minipage}{\dimexpr\textwidth-2\fboxsep-2\fboxrule\relax}
\textbf{Future direction: Prompt-conditioned dynamic timestep selection.}
\end{minipage}%
}

A natural next step is to make the timestep choice dynamic rather than fixed. Different source--target prompt pairs require different noise levels. Appearance-level edits such as changing object color often require a relatively larger noise level, because the edit must affect broad visual regions consistently while still preserving the object identity. In contrast, fine-grained edits such as changing written text, logos, or small local attributes should use a smaller noise level, because the global structure is already correct and the model only needs to adjust high-frequency details. This suggests that future editing systems should infer \lowfreq{$t$} and \highfreq{$t^{**}$} from the relationship between the source and target prompts, instead of using a universal setting for all edit types. More precisely, the model should estimate whether the prompt change asks for broad appearance transformation or precise local refinement, then allocate the noise levels accordingly. Such prompt-conditioned timestep selection could turn the current manually tuned recipe into an adaptive editing policy that better balances target alignment and source preservation across different edit types.

\end{document}